\documentclass{article} 
\usepackage{iclr2022_conference,times}

\usepackage{amsmath,amsfonts,bm}

\def\eqref#1{equation~\ref{#1}}

\def\1{\bm{1}}

\DeclareMathAlphabet{\mathsfit}{\encodingdefault}{\sfdefault}{m}{sl}
\SetMathAlphabet{\mathsfit}{bold}{\encodingdefault}{\sfdefault}{bx}{n}

\newcommand{\E}{\mathbb{E}}

\newcommand{\R}{\mathbb{R}}

\DeclareMathOperator*{\argmin}{arg\,min}

\usepackage{hyperref}
\usepackage{url}
\usepackage{amsmath}
\usepackage{amssymb}
\usepackage{kotex}
\usepackage[pdftex]{graphicx}
\usepackage{multirow}
\usepackage{booktabs}
\usepackage{algorithm}
\usepackage[noend]{algpseudocode}
\usepackage{wrapfig}
\usepackage[font=normalsize,labelfont=bf]{caption}
\usepackage{verbatim}
\usepackage{comment}

\DeclareMathOperator*{\argmaxA}{arg\,max} 

\title{Self-Evolutionary Optimization for Pareto Front Learning}

\author{Simyung Chang, KiYoon Yoo, Jiho Jang \& Nojun Kwak \\
Seoul National University\\
Seoul, Korea \\
\texttt{\{timelighter, 961230, geographic, nojunk\}@snu.ac.kr} \\
}

\iclrfinalcopy 
\begin{document}

\maketitle

\begin{abstract}
Multi-task learning (MTL), which aims to improve performance by learning multiple tasks simultaneously, inherently presents an optimization challenge due to multiple objectives. Hence, multi-objective optimization (MOO) approaches have been proposed for multitasking problems. Recent MOO methods approximate multiple optimal solutions (Pareto front) with a single unified model, which is collectively referred to as \textit{Pareto front learning} (PFL). In this paper, we show that PFL can be re-formulated into another MOO problem with multiple objectives, each of which corresponds to different preference weights for the tasks. We leverage an evolutionary algorithm (EA) to propose a method for PFL called \textit{self-evolutionary optimization} (SEO) by directly maximizing the hypervolume. By using SEO, the neural network learns to approximate the Pareto front conditioned on multiple hyper-parameters that drastically affect the hypervolume. Then, by generating a population of approximations simply by inferencing the network, the hyper-parameters of the network can be optimized by EA. Utilizing SEO for PFL, we also introduce \textit{self-evolutionary Pareto networks} (SEPNet), enabling the unified model to approximate the entire Pareto front set that maximizes the hypervolume. Extensive experimental results confirm that SEPNet can find a better Pareto front than the current state-of-the-art methods while minimizing the increase in model size and training cost.

\end{abstract}

\section{Introduction}
Multi-task learning (MTL) is a learning paradigm, which tries to guide a single shared model to simultaneously learn
multiple tasks while leveraging the domain information included in related tasks as an inductive bias. The goal of MTL is to improve the performance of each task and reduce inference time for conducting all the necessary works for different tasks. 
For some applications, MTL achieves state-of-the-arts performance among large scale neural network-based approaches: computer vision \citep{liu2019end}, natural language processing \citep{liu2019multi} and recommender systems \citep{milojkovic2019multi}. However, it has been observed that since MTL inevitably has multiple objectives corresponding to each task, it is not possible to optimize the performance of all tasks simultaneously due to potential conflicts between these objectives \citep{kendall2018multi}. To  tackle this issue, some attempts have been made to apply multi-objective optimization (MOO) to MTL problems, which optimizes multiple objectives with potential conflicts \citep{sener2018multi, lin2019pareto, ma2020efficient}. MOO aims to find a
Pareto front, which consists of 
a set of solutions that can no longer be optimized without sacrificing the performance of at least one goal, and an element of it is called a Pareto optimal. 
When a preferred direction of the Pareto front, called preference vector (a.k.a. preference ray), is known, it is possible to obtain the corresponding solution \citep{mahapatra2020multi}.

However, in many cases, it is often difficult to predict the trade-off between the actual objectives according to this preference vector, since this trade-off can only be observed once training is complete. Because the number of optimal points in the Pareto front are theoretically infinite, it is very time-consuming to perform training for each optimal point  
to find an appropriate preferred vector. Moreover, for some applications, these preference vectors can change over time and context. A possible way to solve this problem is to approximate the entire set of Pareto front at inference time through a single model. \cite{navon2020learning} call this problem \textit{Pareto front learning} (PFL).

In PFL, a single model is trained to optimize multiple weighted combinations of the original objectives ($\pmb{\mathcal{L}}=(\mathcal{L}_1(\theta), ..., \mathcal{L}_n(\theta))^{T}$) determined by the sampled preference vector ($\pmb{r} \in \R_+^n$). Due to the stochastic nature of the preference vector, the model learns to approximate multiple weighted combinations ($\pmb{\mathcal{L}}^T \pmb{r}^1, \pmb{\mathcal{L}}^T \pmb{r}^2, ...$) during training to ultimately achieve a higher hypervolume (HV)\footnote{Hypervolume can be considered as a measure of distance from a set of optimal points in a Pareto front to the worst possible loss vector called the reference point. For exact definition see \citep{navon2020learning}.}. In this respect, PFL can be re-formulated into another MOO with a goal of maximizing HV, posing new challenges compared to conventional MOO. As is often the case in MOO, we demonstrate that the new objectives, which are determined by the preference vectors, are in conflict with each other in certain circumstances, negatively affecting one another. We call this the \textit{preference conflict.} 

Due to the characteristics of PFL, carefully crafting a sampling procedure or explicitly optimizing the hypervolume is essential. Earlier works in PFL have  implicitly mentioned this issue. \citet{ruchte2021efficient} and \citet{navon2020learning} both need to finetune the parameter of the Dirichlet distribution ($\pmb{\alpha}$). \citeauthor{ruchte2021efficient} additionally add a regularization term that maximizes the cosine similarity between the preference vectors and the original objectives to induce ``spreading" of the Pareto Front, where the coefficient ($\lambda$) needs to be finetuned. In both cases, the tuning of the hyper-parameters ($\lambda$ and $\pmb{\alpha}$) drastically influence the final hypervolume, but the computational cost of grid-searching the values on a validation set is extremely expensive. Moreover, $\pmb{\alpha}$  directly affects the sampling of the preference vectors and $\lambda$ affects the loss term, which are two key components that influence the training trajectory of the model parameters. However, the learning of the model parameters and the grid searching of the hyper-parameters $\pmb{\alpha}$ and $\lambda$ cannot be jointly performed. 

In this paper, we explicitly maximize the hypervolume. First, we discover that what seemed like hyper-parameters can actually be directly optimized to maximize the hypervolume by also conditioning them to the model like the preference vectors. Since using gradient-based optimization for this is non-trivial, we leverage techniques in evolutionary algorithm (EA) to generate, \textit{within the model itself}, population of solutions with different values of $\lambda$ and $\pmb{\alpha}$  without any additional training. To this end, we propose a self-evolutionary optimization for PFL that outperforms all existing MTL methods in a single unified model.

Our contributions are summarized as follows:

(1) We demonstrate that PFL can be interpreted as another MOO problem comprised of conflicting objectives determined by the preference vectors with a goal of maximizing HV. 

(2) We propose self-evolutionary optimization (SEO) for jointly optimizing the hyper-parameters that affect HV along with the model parameters. SEO explores the interval of hyper-parameters by conditioning the model with it during training. Once training is done, this allows the model to generate population of approximated solutions using certain value of hyper-parameters, which can be used for optimization using EA without further training.

(3) We introduce the application of SEO on PFL called Self-evolutionary Pareto Networks (SEPNet). SEPNet is an unified model that can not only approximate the entire set of Pareto front, but can estimate reward of hyper-parameters for various sampling and regularization, allowing direct optimization of HV. Extensive experimental results empirically confirm that the proposed method shows state-of-the-art performance on various MTL tasks and works effectively on large models and large-scale datasets.

\section{Related Works}

\textbf{Multi-task learning (MTL)} 
aims to learn a more general representation by jointly learning on a set of related multiple tasks. By avoiding overfitting on a particular task and taking advantage of the increased data efficiency, a model learned with multiple tasks can potentially achieve better generalization and a faster convergence toward an optimal solution. 
\citet{crawshaw2020multi} categorized MTL into multi-task architecture and multi-task optimization. On the architecture side, \citet{liu2019end} proposed a task-shared feature extractor and task-specific attention modules to produce task-specific features. On the optimization side, \citet{chen2018gradnorm} minimized the difference in the per-task gradients while controlling the weights of per-task loss depending on their learning speed. 

\textbf{Multi-objective optimization}
is used to find a Pareto front.
To populate the Pareto frontier, earlier works \citep{sener2018multi, lin2019pareto} relied on separately learning each point, which is not scalable for large deep networks. Quickly, many efforts have tackled this issue by conditioning on the objective preference during training \citep[PHN, COSMOS]{navon2020learning, ruchte2021efficient} or through transfer-learning on the existing solutions \citep{ma2020efficient}. Linear scalarization (LS), which is a linear weighted sum of the objectives and preference vectors, has been used as a common approach to model a surrogate loss. Since LS often yields solution that does not lie on the preference vector, \citet[EPO]{mahapatra2020multi} propose a gradient-based algorithm to align the loss and the preference vector.
All MOO methods are evaluated on the hypervolume (HV) metric, which is the volume created by the set of pareto optimal points and a reference point. HV considers the quality of the individual Pareto optimal points as well as the diversity of them.
In this work, we re-formulate PFL as another MOO problem and firstly point out a crucial problem of trade-off between the objective preferences due to preference conflict. Moreover, we propose a method to optimize the HV metric directly using evolutionary algorithm.

\textbf{Multi-objective evolutionary alogrithms (MOEA)}
have been one of the most active research areas in EA for recent decades \citep{coello2006evolutionary, deb2011multi}. EAs have been applied to MOO due to their ability to generate sets of solutions called population in a single run.  
The seminal work of \citet{NGSA-III} improves upon one of the early MOEA method \citep{srinivas1994muiltiobjective} by reducing computational complexity and enhancing aspects of genetic algorithm. With the advent of neural networks in MOO, large scale problems have rendered these methods computationally too expensive as they scale poorly to training neural networks. To the best of our knowledge, our proposed method is the first work to use a single neural network through input conditioning to approximate population of multiple parameters to integrate EA into large scale MOO, enabling self-generation of population.

\begin{figure*}[t]
\centering
\includegraphics[width=1.0\linewidth]{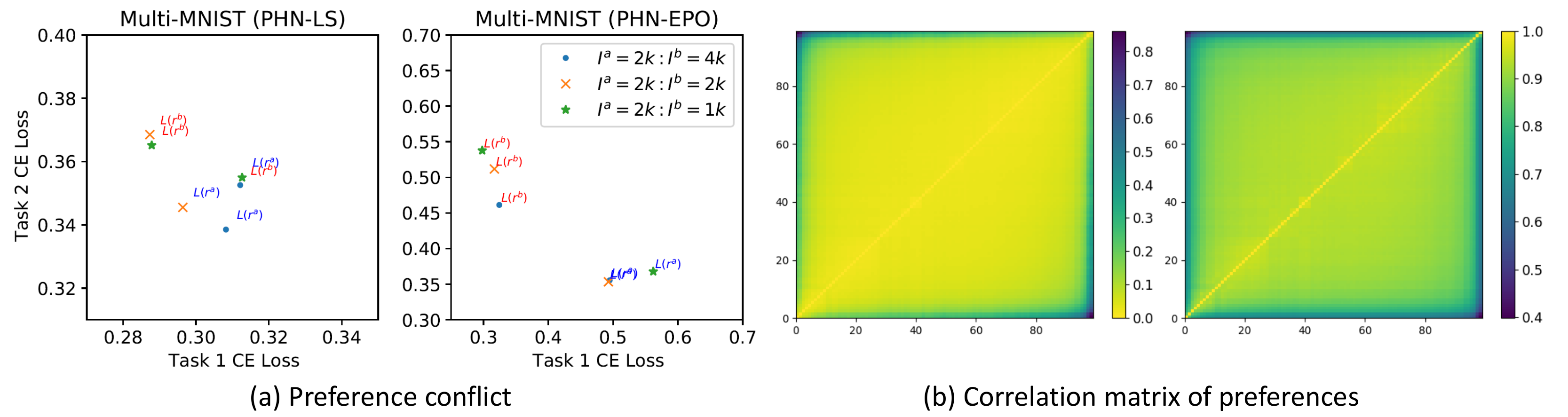}
\vskip -0.05in
\caption{Illustration of preference conflict in PFL (left) and correlation of preferences (right). $I^j$ denotes the number of iterations in which $\pmb{r}^j$ is selected. }
\label{fig:preference}
\end{figure*}

\section{Revisiting Pareto Front Learning }
\label{Sec:revisit_PFL}

A general multi-objective optimization (MOO) problem consisting of $n$ different loss functions can be described by
\begin{equation}
    \pmb{\mathcal{L}}(\theta) = (\mathcal{L}_1(\theta), \mathcal{L}_2(\theta), ..., \mathcal{L}_n(\theta))^{T}, 
\end{equation}
where $\mathcal{L}_i(\theta)$ is the loss for the $i$-th\footnote{In this paper, we use a subscript to represent an element of the corresponding vector, while a superscript denotes its instantiation or a sample.} objective with parameter $\theta$. MOO exploits the shared structures and information among them to optimize all objectives simultaneously. However, no single solution can optimize all objectives at the same time. Instead, we expect to obtain a solution which cannot improve the corresponding objective without degradation in another.
Following the definition of \cite{navon2020learning}, $\theta^1$ is said to dominate $\theta^2$ if $\mathcal{L}(\theta^1) \preceq \mathcal{L}(\theta^2)$ and $\mathcal{L}_i(\theta^1) < \mathcal{L}_i(\theta^2)$ for some $ i \in [n]$. An optimal solution that is not dominated by any other point is called \textit{Pareto Optimal}.

\textit{Pareto Front Learning} (PFL) is an approach to approximate the \textit{Pareto front}, the set of all Pareto optimal, at inference time through a single model \citep{navon2020learning}. Existing works minimize the empirical risk of the following loss
\begin{equation}
    \mathcal{\bar{L}}_{PFL}(\theta) = \E_{\pmb{r} \sim Dir(\pmb{\alpha})}\sum_i g(r_i,\mathcal{L}_i(\theta)) = \E_{\pmb{r} \sim Dir(\pmb{\alpha})}\sum_i r_i \mathcal{L}_i(\theta)
\label{eq:standard PFL}
\end{equation}

where $g()$ is a function that weights each objective $\mathcal{L}_i$ by the corresponding weight $r_i$ which is sampled from an $n$-dimensional Dirichlet distribution. The last equality comes from using LS as $g()$.

\subsection{Preference Conflict}
\label{sec:conflict}
In this secsion, we describe the additional challenges of PFL that arise when a single model approximates multiple optimal points.
We can interpret PFL as approximating distinct $m$ points of the pareto front denoted as $\mathcal{L}(\theta, \pmb{r}^j), j \in [m]$, with a single network: 
\begin{equation}
    \pmb{\mathcal{L}}_{PFL}(\theta) = (\mathcal{L}(\theta, \pmb{r}^1), \mathcal{L}(\theta, \pmb{r}^2), ..., \mathcal{L}(\theta, \pmb{r}^m))^T.
\label{eq:PFL}
\end{equation}
This give rises to another MOO problem for PFL. Here, we analyze the characteristics of PFL with distinct preference vectors  $\pmb{r}^j$. 

Figure \ref{fig:preference}(a) 
shows the results of using two preference vectors $\pmb{r}^a$ and $\pmb{r}^b$ and varying their sampling frequency for PHN \citep{navon2020learning}. To show how each affects one another, we fix the sampling frequency of $\pmb{r}^a$ and increase or decrease that of $\pmb{r}^b$. The results indicate that this affects not only the results of $\pmb{r}^b$, but also influences that of $\pmb{r}^a$ as well, indicating the existence of conflictual interaction between the two. We define this as the \textit{preference conflict}. 
 
Figure \ref{fig:preference} 
shows that the preference conflict also exists when EPO, instead of LS, is used for optimization. 
Due to this conflict, one may not be able to improve the performance of one preference objective without sacrificing another preference objective. 
In Figure \ref{fig:preference}(b), we show the correlation between individually trained models for 100 linearly spaced preference vectors from 0.01 to 0.99. This is computed by the average of the JS divergence (left) and cosine similarity (right) of the outputs of each models. The diagonal elements indicate perfect correlation and all the preference vectors have some degree of correlation with its adjacent elements as shown by the high value of cosine similarity and low JS divergence. 
Note the gradation of the correlation between adjacent vectors slowly decreases (higher JS divergence, lower cosine similarity) as we move towards both ends of the preference vectors.  
Thus, solving the new MOO problem involves finding the optimal point considering the trade-off between the highly correlated preference vectors. We offer how to mitigate this problem in Section \ref{section4} by introducing the sparse sampling strategy.

\begin{figure}
  \begin{minipage}{\textwidth}
  \begin{minipage}[c]{0.52\textwidth}
    \centering
    \includegraphics[width=1.0\linewidth]{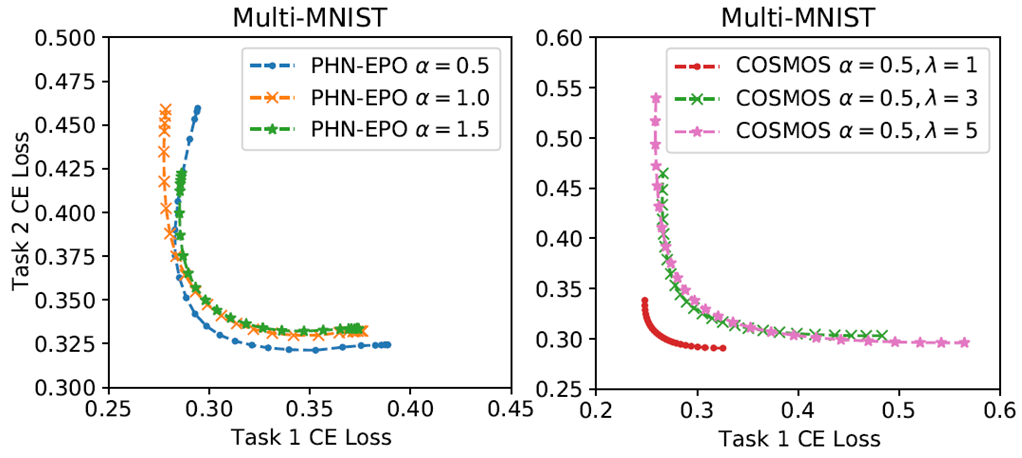}
    \vskip -0.1in
    \captionof{figure}{Comparison of the Pareto fronts on different hyper-parameters. For both methods, the Pareto front and hence the HV are sensitive to the hyper-parameters.}
    \label{fig:conflict}
  \end{minipage}
  \hfill
  \begin{minipage}[c]{0.46\textwidth}
    \captionof{table}{Hypervolume on Multi-MNIST depending on hyper-parameter ${\alpha}$ and $\lambda$}
    \centering
    \resizebox{1.\linewidth}{!}{
    \begin{tabular}{lccc}
    \toprule
    Method & $\alpha = 0.5$ & $\alpha = 1.0$ & $\alpha = 1.5$\\
    \midrule     
    PHN-LS & 2.84 & 2.84 & 2.86 \\
    PHN-EPO & 2.88 & 2.87 & 2.86 \\
    COSMOS ($\lambda=1$) & 2.99 & 2.95 & 2.97 \\
    COSMOS ($\lambda=3$) & 2.94 & 2.95 & 2.95 \\
    COSMOS ($\lambda=5$) & 2.96 & 2.97 & 2.95 \\
    \bottomrule
    \end{tabular}
    }
    \end{minipage}
  \end{minipage}
\end{figure}

\subsection{Hypervolume Maximization}
\label{Sec:hv max.}
Here we motivate SEO from our new formulation (Eq. \ref{eq:PFL}).
Linear scalarization (LS) is the most straightforward approximation of multiple objectives in MOO. 
LS defines a single surrogate loss $\mathcal{L}_{\pmb{r}}(\theta) = \sum_i r_i \mathcal{L}_i(\theta)$ given a preference vector $\pmb{r} \in \R_+^n$ with $\sum_{i} r_i = 1$ for multiple objectives.
In this paper, we define a new surrogate loss 
\begin{equation}
\mathcal{L}'_{\pmb{r}}(\theta) = \frac{1}{m} \sum_{i}^m \mathcal{L}(\theta, \pmb{r}^i)   
\end{equation}
and show that $\mathcal{L}'_{\pmb{r}}$ is exactly the same as $\mathcal{L}_{\pmb{r}}$ in expectation. 
The proof is quite straightforward as follows:
As in previous works, if we sample preference vectors according to the Dirichlet distribution with parameter $\pmb{\alpha}$, the expected loss becomes
\begin{equation}
    \begin{split}
    \mathcal{\bar{L}}_{PFL}'(\theta) &= \E_{\{\pmb{r}^1, \cdots,\pmb{r}^m\}  \sim Dir(\pmb{\alpha})} \frac{1}{m} \sum_{i}^m \mathcal{L}(\theta, \pmb{r}^i) \\
    & = \E_{\pmb{r} \sim Dir(\pmb{\alpha})} \mathcal{L}(\theta, \pmb{r}) = \E_{\pmb{r} \sim Dir(\pmb{\alpha})} \sum_j^n r_j \mathcal{L}_j(\theta),
    \end{split}
\label{eq:standard PFL2}
\end{equation}
which is identical to Eq. \ref{eq:standard PFL}. Note that the expected loss $\mathcal{\bar{L}}_{PFL}$ depends on hyper-parameter $\pmb{\alpha}$.

Since the final goal of PFL is to maximize the hypervolume (HV), we seek to solve  $\argmaxA_{\theta,\pmb{\alpha}}HV(\theta, \pmb{\alpha})$.
For COSMOS, an additional hyper-parameter $\lambda$ exists for maximizing the cosine similarity of the preference vector $\pmb{r}$ and objective vector $\pmb{\mathcal{L}}$, which leads to finding the optimal model parameters by  $\theta^* = \argmin_{\theta}\E_{\pmb{r} \sim Dir(\pmb{\alpha})}\sum_i r_i \mathcal{L}_i(\theta)-\lambda C(\pmb{r}, \pmb{\mathcal{L}})$ where $C$ denotes the cosine similarity between two vectors.
Note that to reduce the search space, we set all the elements of $\pmb{\alpha}$ identically as $\alpha$.
By combining all the hyper-parameters as $\phi=[{\alpha}, \lambda]^T$, our PFL problem ultimately solves 
\begin{equation}
    (\theta^*, \phi^*) = \argmaxA_{\theta,\phi} HV(\theta, \phi).
    \label{eq:HV_max}
\end{equation}
We show in Table 1 and Figure \ref{fig:conflict} that $\alpha$ and $\lambda$ drastically influences HV and the final Pareto front depending on the value, which necessitates grid-search. 

\section{Proposed Method}
\label{section4}
\begin{algorithm}[t]
\begin{algorithmic}
\State \textbf{Input:} noise standard deviation $\sigma$, population size $n$
\For {\textit{training epochs}: $k=0, 1, 2, ...$}
    \For {\textit{training iterations}: $t=0, 1, 2, ...$}
        \State Sample $\epsilon \sim U(l, h)$
        \State Sample $(x,y) \sim p_D$
        \State $g_\theta \leftarrow \nabla_{\theta}\mathcal{L}(y, f(x, \phi+\epsilon;\theta), \phi+\epsilon)$
        \State $\theta \leftarrow \theta - \gamma_{\theta}g_\theta $
    \EndFor
    \State Sample $\epsilon_1, ..., \epsilon_n \sim \mathcal{N}(0,I)$
    \State Compute $\mathcal{S}_i =\mathcal{S}(f(\phi+ \sigma\epsilon_i;\theta))$ for $i \in [n]$ on the entire validation dataset
    \State $\phi \leftarrow \phi + \gamma\frac{1}{n\sigma}\sum_{i=1}^{n}\mathcal{S}_i\epsilon_i$
\EndFor
\end{algorithmic}
\caption{Self-evolutionary optimization (SEO)}
\label{alg:SEO}
\end{algorithm}

\subsection{Self-evolutionary Optimization}
\label{sec:SEO}
As mentioned in Section \ref{Sec:revisit_PFL}, PFL tries to find a network with optimal model parameters $\theta$ and hyper-parameters $\phi$. However, training a neural network from scratch to find $\phi$ is computationally very expensive.
Instead, we design our network to approximate the HV when certain $\phi$ values are used in inference by conditioning them while training. Then, the approximated population of solutions can be optimized by an algorithm based on evolutionary strategy (ES) to find the optimal $\phi$. 
We call our method self-evolutionary optimization (SEO). 

Given a fitness function $\mathcal{S}$\footnote{Here, the hypervolume is directly used as our fitness function.}, hyper-parameters $\phi$, a noise standard deviation $\sigma$, the ES algorithm that uses Gaussian search makes use of an estimator for the gradient $\nabla_{\phi} \E_{\epsilon \sim N(0,I)} \mathcal{S}(\phi + \sigma\epsilon) $ \citep{nesterov2017random, salimans2017evolution}. 
Our SEO that utilizes this ES algorithm consists of two phases.
First, we redefine our original network $f(x;\theta)$ as a conditional function of $\phi$ as well by $f(x, \phi;\theta)$. In the training phase, $f(x, \phi;\theta)$ is trained through gradient descent with the following gradient: 
\begin{equation}
\nabla_{\theta}\E_{{\epsilon \sim U(l,h)\\(x,y) \sim P_D}} \mathcal{L}(y, f(x, \phi+\epsilon;\theta), \phi+\epsilon).
\label{eq:obj_theta}
\end{equation}
Here, $U(l,h)$ indicates uniform sampling between $l$ and $h$\footnote{A uniform distribution is used in the training phase as we find the Gaussian distribution deteriorates the performance by excessively sampling values near the mode.}.

Then, in the validation phase, we search optimal $\phi$ using the ES algorithm. Using $n$ population of $\phi$, which is obtained using additive Gaussian noise, the score according to each $\phi$ is estimated through $f(\phi;\theta)$ and the fitness function $\mathcal{S}$. We directly optimize $\phi$ with stochastic gradient ascent using the following gradient estimator on the right \citep{salimans2017evolution}: 
\begin{equation}
\nabla_{\phi}\E_{\epsilon \sim N(0, I) }\mathcal{S}(f(\phi+ \sigma\epsilon;\theta)) = \frac{1}{\sigma} \E_{\epsilon \sim N(0, I) }\{\mathcal{S}(f(\phi+ \sigma\epsilon;\theta)) \epsilon\}.
\label{eq:obj_phi}
\end{equation}

Algorithm \ref{alg:SEO} illustrates this process of alternating between the training phase and the validation phase.

\subsection{Sparse Sampling}
\label{sparse sampling}
We showed in Section \ref{sec:conflict} that PFL learns a single network to solve MOO of highly correlated preference vectors. These vectors are sampled from a continuous distribution and used as additional input for the network,  meaning the model has to theoretically approximate infinite objectives.
Moreover, due to the high correlation between the adjacent samples, continuous sampling may interfere with the weights of adjacent samples.  
To reduce the overhead caused by additional infinite inputs and the influence of adjacent preference vectors, we propose a simple trick called ``\textit{sparse sampling}". Sparse sampling quantizes the samples from a continuous distribution to a smaller set of discrete finite values.
We define a quantization function $Q(x;\Gamma,\tau)$ that quantizes $x$ to $\overline{x}$, which corresponds to the center of a bin among uniformly divided $\tau$ number of bins for range $\Gamma=(l,h)$.
Applying the quantization function to $\pmb{r}$ for $\E_{\pmb{r} \sim Dir(\alpha)}\sum_i r_i\mathcal{L}_i(\theta)$ yields $\E_{\pmb{r} \sim Dir(\alpha)}\sum_i \overline{r_i}\mathcal{L}_i(\theta)$ .

\subsection{Self-evolutionary Pareto Network}
Here, we introduce our method by applying SEO to Pareto front learning called \textit{self-evolutionary Pareto networks} (SEPNet). SEPNet defines the fitness function $\mathcal{S}$ as HV (Eq. \ref{eq:HV_max}) and optimizes it via SEO. SEPNet first optimizes the model parameter $\theta$ while being conditioned on $\phi$, which consists of $\alpha$ and $\lambda$, i.e, $\phi = [\alpha, \lambda]^T$. And we apply sparse sampling for preference rays $r$ and $\phi$.
In this case, the gradient in Eq. \ref{eq:obj_theta} becomes 
\begin{equation}
\nabla_{\theta}\E_{{\epsilon \sim U(l,h), \pmb{r} \sim Dir(\phi_{\alpha}+\epsilon_{\alpha}), (x,y) \sim P_D}} \mathcal{L}(y, SEPNet(x, \overline{\pmb{r}}, \overline{\phi+\epsilon};\theta), \overline{\phi+\epsilon}).
\label{eq:obj_SEPNet}
\end{equation}

Then in the validation phase, it optimizes $\phi$ using Eq. \ref{eq:obj_phi}.
To condition $\pmb{r}$ and $\phi$ to our network, we utilize a condition injection module that takes in the input condition $C=[r, \phi]^T$ (i.e, instantiation of $\pmb{r}$ and $\phi$). Using conditioning techniques from \citet{huang2017arbitrary, perez2018film}, 
we define the conditioning module $G$ as an affine transformation function on the feature space $x$ as $\tilde{x} = G_{scale}(C)x + G_{shift}(C)$, where $G$ is a MLP with two fully connected layers. And two or three conditioning modules are inserted after different layers of the baseline networks. Therefore, the increase in parameter is negligible.

\section{Experiments}
In this section, we verify our SEPNet advances the state-of-the-art in PFL on various datasets, significantly outperforming existing works. Additionally, we conduct an ablation study to test the effectiveness of self-evolutionary optimization and sparse sampling.

\begin{figure*}[t]
\centering
\includegraphics[width=0.95\linewidth]{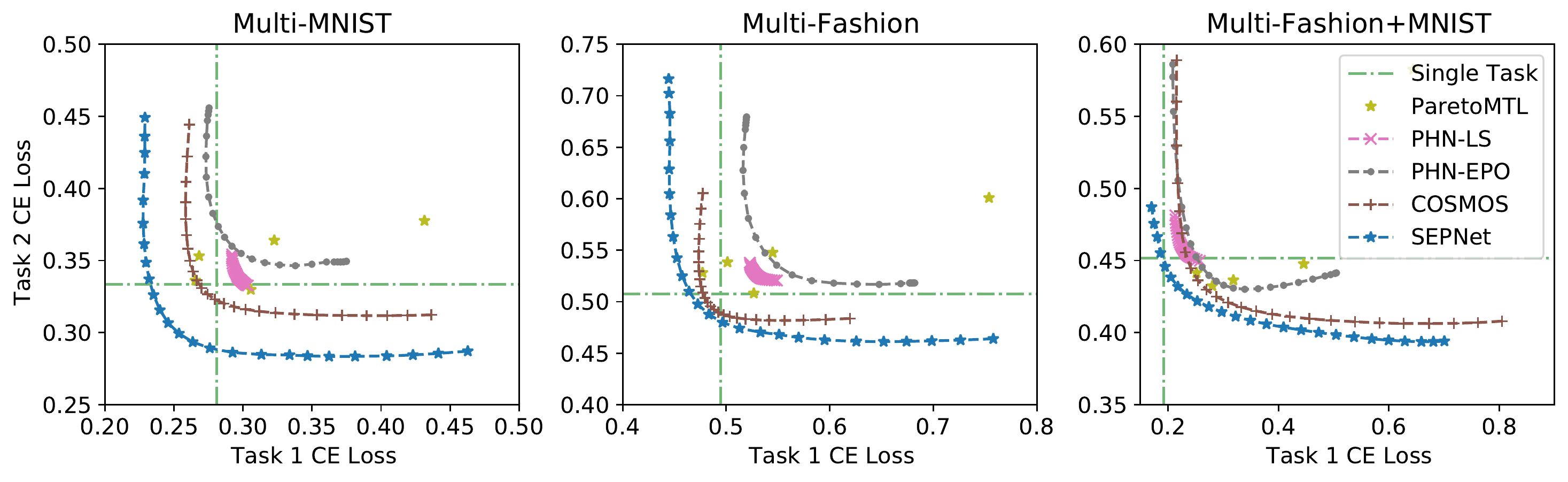}
\includegraphics[width=0.95\linewidth]{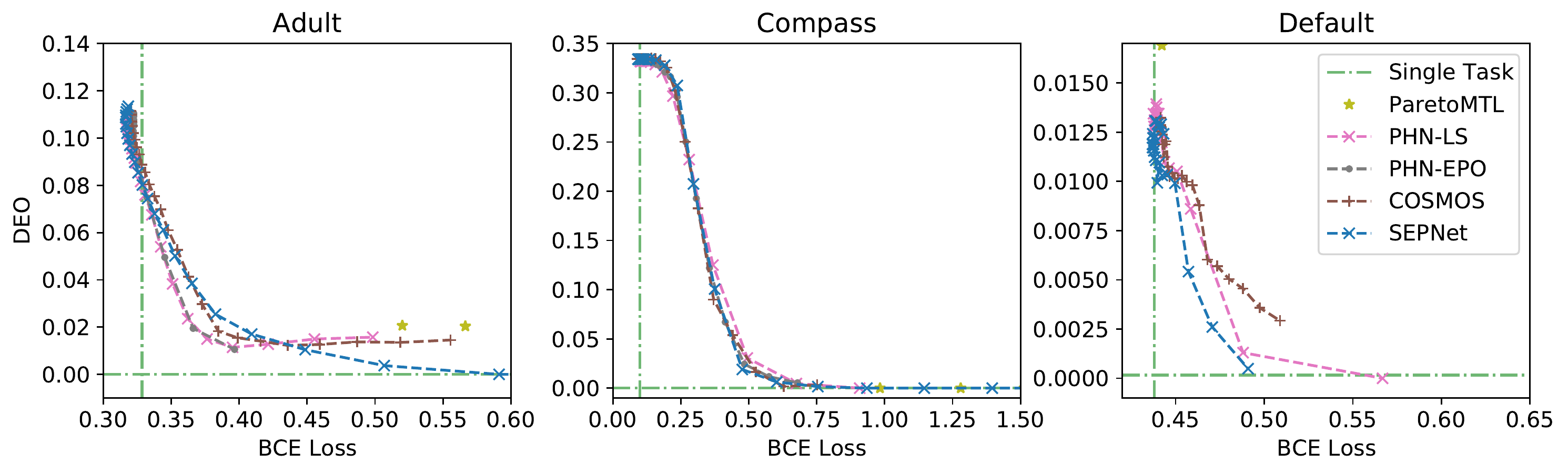}
\vskip -0.1in
\caption{Comparison of the Pareto fronts on Image classification (top) and Fairness (bottom) tasks. }
\label{fig:Multi}
\end{figure*}

\begin{table}[]
\caption{Quantitative Evaluation on Image Classification and Fairness. }
\centering
\resizebox{1\linewidth}{!}{
\begin{tabular}{l|ccc|ccc}
\toprule
Method &Multi-MNIST & Multi-Fashion & Multi-F+MNIST &  Adult & Compass & Default \\
\midrule
PMTL    & 2.90   & 2.27   & 2.74   & 2.92   & 2.15   & 3.10  \\
\midrule
PHN-EPO & 2.86   & 2.20   & 2.80   & 3.34   & 3.71   & 3.11  \\
PHN-LS  & 2.85   & 2.19   & 2.77   & 3.34   & 3.71   & 3.12  \\
COSMOS  & 2.94   & 2.32   & 2.84   & 3.34   & \textbf{3.72}   & 3.12  \\
\midrule
\textbf{SEPNet} (ours) &\textbf{3.04}   & \textbf{2.39}   & \textbf{2.93}    & \textbf{3.36}   & \textbf{3.72}   & \textbf{3.13}   \\
\bottomrule 
\end{tabular}
\label{table:Image Classification}
}
\end{table}

\subsection{Experimental Settings}
Our experiments are conducted following the general settings of recent MOO works \citep{lin2019pareto, navon2020learning}. 
We compare with \citet[PHN-LS, PHN-EPO]{navon2020learning}, \citet[ParetoMTL]{lin2019pareto} and \citet[COSMOS]{ruchte2021efficient} as baselines which generate Pareto fronts with gradient-based method. The baselines are trained with officially implemented code,
and early stopping is applied using the HV on the validation set.
Unless otherwise stated, we use Adam \citep{kingma2014adam}, a batch size of 256 and a learning rate 1e-3. The reported scores and Pareto fronts are the average of learning 5 times each, and we take (2,2) as a reference point of HV.

For our SEO, we use $\sigma=0.1$ and population size $n=10$ for the ES algorithm, and fix the learning rate to 0.01. We alternate between $\alpha$ and $\lambda$ every 5 epochs for optimization. We evaluate the HV of every population using 5 preference rays on the entire validation dataset. We reduce the number of preference rays from 25 to 5 in the validation phase, but the overall validation time is approximately doubled by using $n=10$. 
The overhead in training time was within 10\% compared to a single objective baseline.
For sparse sampling, we quantize the samples of Dirichlet distribution by a grid of 0.1. For SEO, we choose the offset of $\alpha$ to be between -0.2 and +0.2, which is sampled within a discrete bin of size 0.1. Similarly, the offset of $\lambda$ is between -1.0 and 1.0 with bin size 0.5. SEO optimizes $\alpha\text{ and }\lambda$ in the validation set and the found values are used for evaluation in the test set. 

\textbf{Image Classification} We evaluate our method on multi-MNIST \citep{sabour2017dynamic} and two variants of the FashionMNIST dataset \citep{xiao2017fashion} called multi-Fashion and Fashion+MNIST  \citep{lin2019pareto}. For each data set, two images are randomly sampled from the corresponding dataset (e.g. MNIST + MNIST, Fashion + MNSIT) , then the images are slightly overlapped by placing them on top-left (TL) and bottom-right (BR). In MTL, the model classifies both instances at the same time. Each dataset consists of 120,000 training examples and 20,000 test examples, we allocate 10\% of training examples for validation. We use LeNet \citep{lecun1999object} and learning rate are decayed by 0.1 at 40, 80, 90 epoch.

\textbf{Fairness} We conduct experiments for incorporating fairness into the model. We use the Adult \citep{Dua:2019}, Compass \citep{angwin2016machine} and Default \citep{yeh2009comparisons} dataset. We choose `sex' as a binary sensible attribute and
optimize the fairness objective which is a hyperbolic tangent relaxation of \textit{difference of equality of opportunity} ($\widehat{DEO}$) defined in \cite{padh2020addressing}.
We train two hidden layers (60 and 25 dimensions) of MLP with ReLU activation for 50 epochs. Each dataset is divided into train/validation/test sets of 70\%/10\%/20\%, respectively.

\textbf{CelebA} CelebA \citep{liu2015deep} is a large-scale face attributes dataset with more than 200K celebrity images. We adopt the experimental setup of COSMOS. 
We rescale the images of CelebA to 64x64, and adopt ImageNet-pretrained EfficientNet-B4 \citep{tan2019efficientnet} as a backbone. SEPNet is trained by setting the learning rate to 5e-4 and the batch size to 32. For MTL, we optimize binary cross-entropy for two easy and two hard tasks as done by \cite{ruchte2021efficient}. Facial attributes such as Oval Face and Pointy Nose are combined to make the hard task, and Goatee and Mustache are used as the easy task.

\begin{figure}
  \begin{minipage}{\textwidth}
  \begin{minipage}[c]{0.47\textwidth}
    \centering
    {\includegraphics[width=1.0\linewidth]{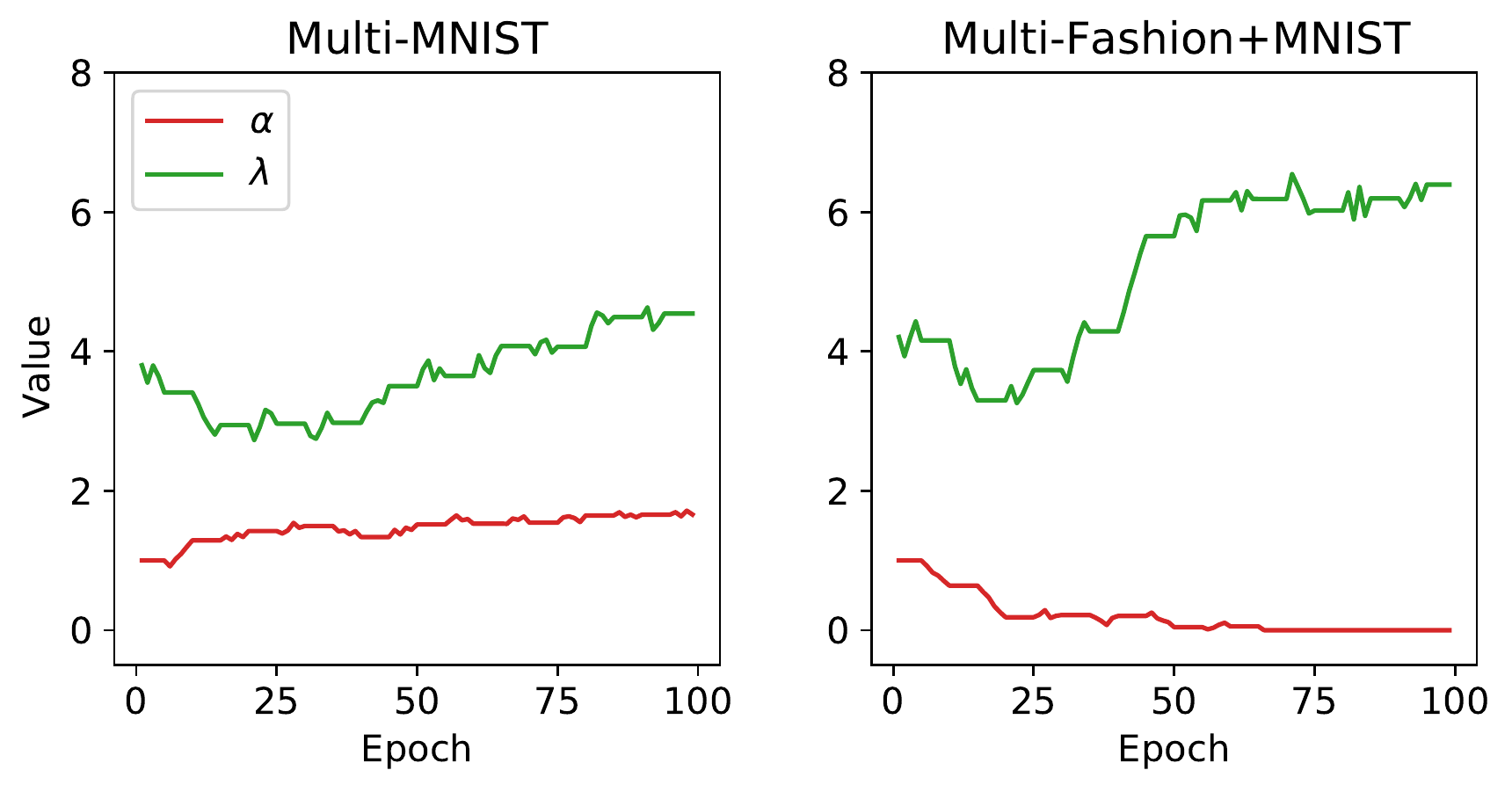}}
    \vskip -0.1in
    \caption{Trends in $\alpha$ and $\lambda$ during training for the two Image Classification datasets.}
    \label{fig:SEO_alpha}
  \end{minipage}
  \hfill
  \begin{minipage}[c]{0.50\textwidth}
    \centering
    {\includegraphics[width=1.0\linewidth]{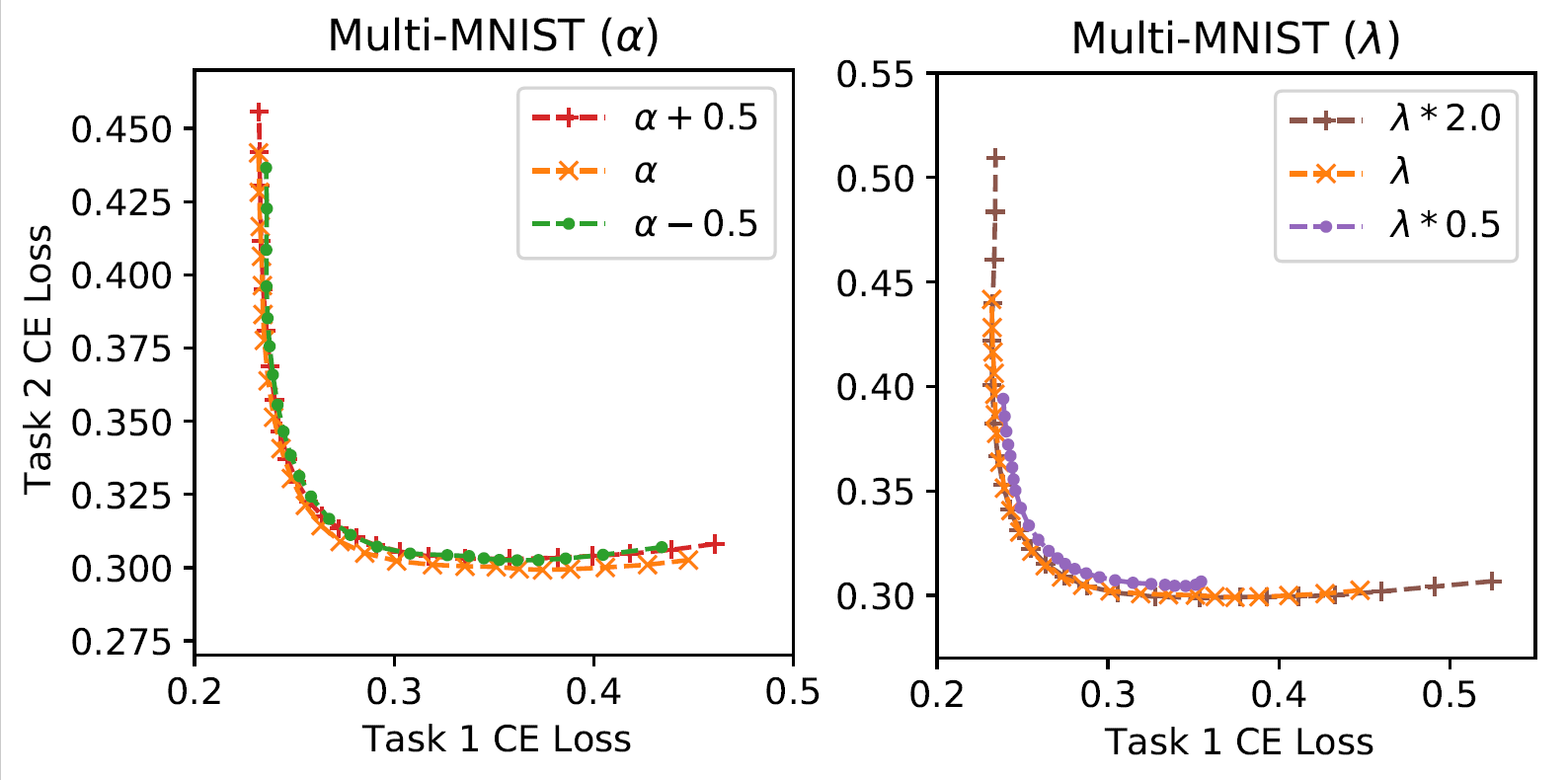}}
    \vskip -0.1in
    \caption{Various generated Pareto fronts by conditioning the network inputs $\alpha$ and $\lambda$. We see that by SEO, the network is able to generate various Pareto fronts dependent on the hyper-parameters.}
    \label{fig:SEO_shift}
    \end{minipage}
  \end{minipage}
\end{figure}  

\begin{figure*}[t]
\centering
{\includegraphics[width=0.9\linewidth]{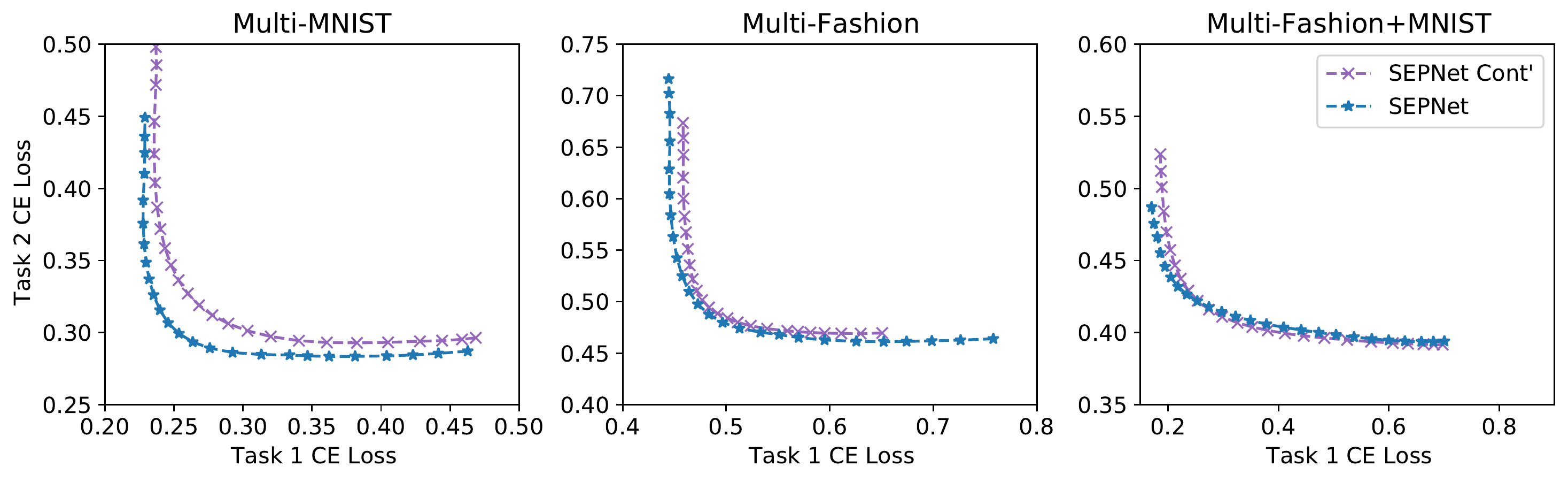}}
{\includegraphics[width=0.9\linewidth]{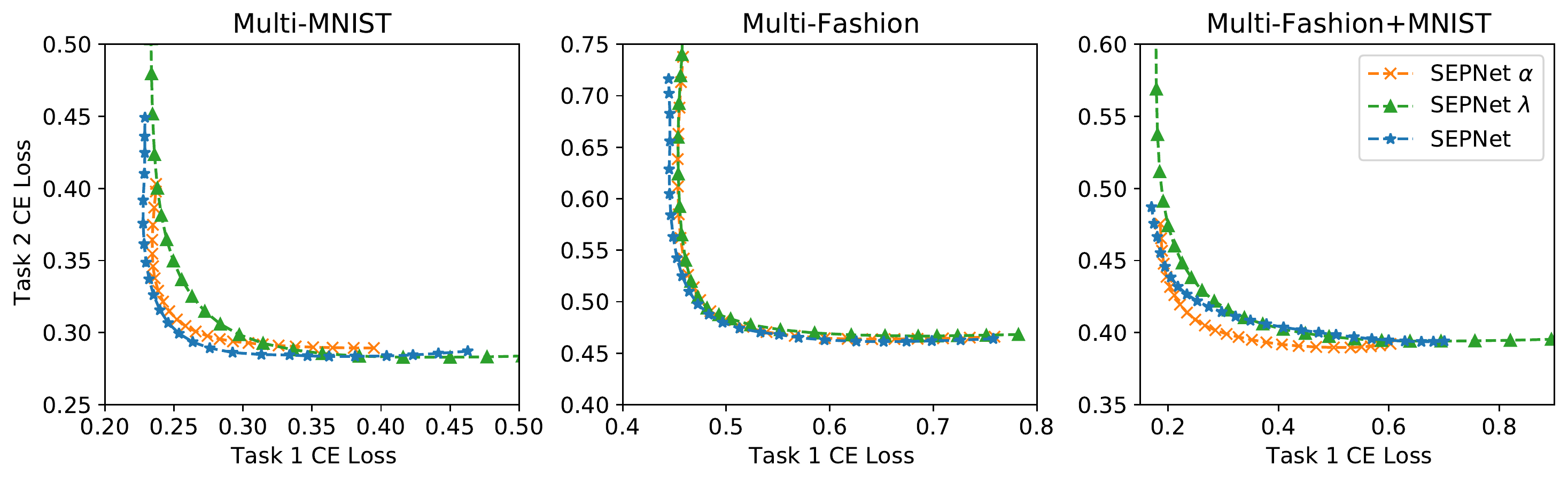}}
\vskip -0.1in
\caption{Qualitative results of ablation study. The first row illustrates the effect of  sparse sampling, while the second row shows that of SEO.  
}
\label{fig:ablation}
\end{figure*}

\subsection{Results and Discussion}

Table \ref{table:Image Classification} shows that SEPNet outperforms both methods using a seperate model for each preference vector (PMTL) and methods using an unified model (PHN, COSOMOS). 
Qualitatively, Figure \ref{fig:Multi} shows that SEPNet creates a wider Pareto front with superior solutions.  
In particular, for the multi-FASION+MNIST dataset, SEPNet attains solutions that perform better than that of single task for Task 1, whereas all the other methods are unable to reach this point. 
 
On the Fairness dataset, we quantitatively attain higher HV than the baseline, but the improvement is less noticeable. We believe this is due to the use of small neural network (e.g. 2-layer MLP) that has a limited capacity for learning the whole Pareto frontier. 

Figure \ref{fig:SEO_alpha} shows the trend of $\alpha\text{ and }\lambda$ while training for Image Classification. 
Note that the different optimal values are found in each dataset, which implies that to find the optimal values for other methods, these should be tuned as hyper-parameters for each dataset by grid-search.
After training, the two parameters can be conditioned to different values for inference to generate multiple Pareto fronts as shown by Figure \ref{fig:SEO_shift}. Hence by following the protocol explained in Section \ref{sec:SEO}, we can find the optimal parameters in the validation set through self-evolutionary optimization.

\subsection{Ablation}
\textbf{Sparse Sampling}
Figure \ref{fig:ablation} indicates the Pareto front of SEPNet Cont', which uses continuous sampling instead of sparse sampling. For all three datasets, SEPNet creates superior optimal points overall.  
Also note that while sparse training only samples 10 points from the segment $[0,1]$, it is able to interpolate all the 25 preference vectors, which were not seen during training.

\textbf{Optimization Parameters using SEO}
In the second row of Figure \ref{fig:ablation}, ``SEPNet $\alpha$" only optimizes $\alpha$ while fixing $\lambda$, and vice-versa. We find that the optimizing the two parameters has a distinct effect. 
For instance, optimizing for $\lambda$ leads to a wider Pareto front, which offers more diverse solutions and increases HV. On the other hand, optimizing for $\alpha$ usually advances the preference vectors near the center ray (0.5, 0.5). This phenomenon is most noticeable for Multi-MNIST. Overall, though training for only single parameter may have individual preference vectors that perform better than that of SEPNet that optimizes both, SEPNet achieves the highest HV. Quantitative analysis can be found in Appendix.

\begin{figure}
  \begin{minipage}{\textwidth}
  \begin{minipage}[c]{0.55\textwidth}
    \centering
    \includegraphics[width=1.0\linewidth]{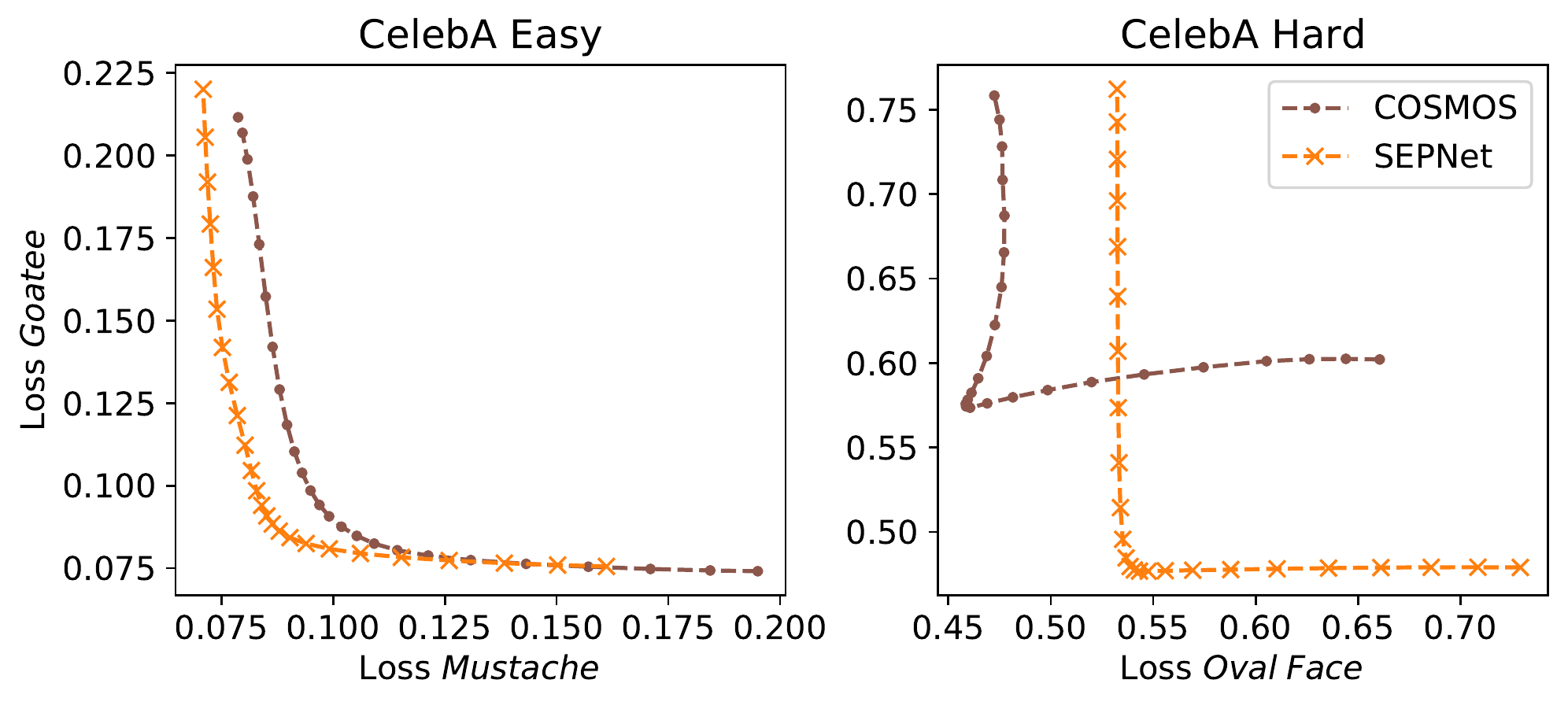}
    \vskip -0.1in
    \captionof{figure}{The Pareto fronts on CelebA}
    \label{fig:celeba}
  \end{minipage}
  \hfill
  \begin{minipage}[c]{0.44\textwidth}
    \captionof{table}{Hypervolume on CelebA Easy and Hard tasks}
    \label{table:celeba}
    \centering
    \resizebox{1.\linewidth}{!}{
    \begin{tabular}{lcc}
    \toprule
    Method & CelebA Easy & CelebA Hard\\
    \midrule
    Single Task & 3.719 & 2.222 \\
    COSMOS & 3.706 & 2.221 \\
    SEPNet & \textbf{3.713} & \textbf{2.235} \\
    \bottomrule
    \end{tabular}
    }
    \end{minipage}
  \end{minipage}
\end{figure}  

\subsection{Scaling Up SEPNet}
We experiment with CelebA to verify our SEPNet can also be applied to large dataset, using a modern architecture model (i.e., EfficientNet-B4).
Because PHN and ParetoMTL have poor scalability, we compare our method with COSMOS and Single Task baseline, trained using a single task objective. Table \ref{table:celeba} shows the quantitative comparison of SEPNet and other baselines. Since a Single Task does not approximate the Pareto front for multiple reference rays, the HV is calculated at the cross point of the two task losses following convention.  COSMOS did not show better results than Single Task baseline in both Easy and Hard tasks. In contrast, SEPNet has higher HV than all baselines for Hard Task and obtains similar results to Single Task baseline on Easy Task. And as shown in Figure \ref{fig:celeba}, our method approximates a well-spread smooth Pareto front even in CelebA. These results indicate that SEPNet can approximate the entire Pareto fronts on a large dataset using a modern architecture.

\section{Conclusion}
We propose a new perspective to PFL as another MOO problem. We tackle conflicts between the preference vectors and propose to optimize hyper-parameters jointly with model parameters by leveraging an evolutionary strategy. 
In doing so, we set the fitness function for hyper-parameter optimization as the hypervolume and could achieve SOTA performances on various MTL tasks. Our work can be improved upon by utilizing a more effective evolutionary algorithm such as genetic algorithms. In addition, additional methods can be devised to optimize the model parameters also for the hypervolume metric. 

\clearpage
\bibliography{iclr2022_conference}
\bibliographystyle{iclr2022_conference}

\clearpage
\appendix
\section{Appendix}

\setcounter{table}{0}
\renewcommand{\thetable}{A\arabic{table}}
\setcounter{figure}{0}
\renewcommand{\thefigure}{A\arabic{figure}}

\subsection{Experiment Details}

\subsubsection{Experimental Details of Preference Conflict}
For each training, preference ray $\pmb{r}^a$ is used for $I^a$ iterations, and  $\pmb{r}^b$ is used for $I^b$ iterations. $\pmb{r}^a$ or $\pmb{r}^b$ is chosen randomly according to each number of iterations.

\begin{table}[h]
\centering
    \begin{tabular}{l}
    \toprule
    $\pmb{r}^a$ = [0.2, 0.8] \\
    $\pmb{r}^b$ = [0.8, 0.2] \\
    \bottomrule
    \end{tabular}
\end{table}

\subsubsection{Position of Conditioning Modules}
We use an MLP with two fully connected (FC) layers as a conditioning module. The first FC layer has the same number of nodes as the target feature layer and the second FC layer is twice the first. We utilize the half of the output as a scaling parameter and the other half as a shifting parameter.

\begin{figure}[h]
  \begin{minipage}{\textwidth}
  \begin{minipage}[c]{0.47\textwidth}
\caption{ LeNet (Method \& Multi-MNIST \& Multi-Fashion \& Multi-F+MNIST)}
\centering
    \begin{tabular}{c}
    \toprule
    \textbf{Input} $x$ \\
    \midrule
    Convolution \\
    \midrule
    \textbf{Conditioning Module} \\
    \midrule
    Convolution \\
    \midrule
    \textbf{Conditioning Module} \\
    \midrule
    Shared Fully Connected \\
    \midrule
    Head Fully Connected \\
    \bottomrule
    \end{tabular}   
  \end{minipage}
  \hfill
  \begin{minipage}[c]{0.47\textwidth}
\caption{ MLP (Adult \& Compass \& Default)}
\centering
    \begin{tabular}{c}
    \toprule
    \textbf{Input} $x$ \\
    \midrule    
    Fully Connected \\
    \midrule
    \textbf{Conditioning Module} \\
    \midrule
    Fully Connected \\
    \midrule
    \textbf{Conditioning Module} \\
    \midrule
    Head Fully Connected \\
    \bottomrule
    \end{tabular}    
    \end{minipage}
  \end{minipage}
\end{figure}

\begin{table}[h]
\caption{
EfficientNet-B4 (CelebA)
}
\centering
    \begin{tabular}{c}
    \toprule
    \textbf{Input} $x$ \\
    \midrule    
    Stem Convolution \\
    \midrule
    MBConvBlocks (0-10)\\
    \midrule
    \textbf{Conditioning Module} \\
    \midrule
    MBConvBlocks (11-20) \\
    \midrule
    \textbf{Conditioning Module} \\
    \midrule
    MBConvBlocks (21-30)\\
    \midrule
    \textbf{Conditioning Module} \\
    \midrule
    MBConvBlocks (31)\\
    \midrule
    Head Convolution\\
    \midrule    
    Head Fully Connected \\
    \bottomrule
    \end{tabular}
\end{table}

\clearpage
\subsection{Additional Experiments}

\begin{table}[h]
\caption{
\textbf{Hypervolume with or without Sparse Sampling}  \textit{SEPNet Cont'} denotes the SEPNet without Sparse Sampling.
}
    \vskip 0.05in
    \label{table:discrete}
    \centering
    \begin{tabular}{lccc}
    \toprule
    Method & Multi-MNIST & Multi-Fashion & Multi-F+MNIST\\
    \midrule     
    \textbf{SEPNet} & \textbf{3.04} & \textbf{2.39} & \textbf{2.93}\\
    SEPNet Cont' & 3.01 & 2.36 & 2.91\\
    \bottomrule
    \end{tabular}
\label{table:exp_sparse}
\end{table}

\begin{table}[h]
\caption{
\textbf{Hypervolume according to a selection of hyper-parameters for SEO}
}
    \vskip 0.05in
    \label{table:discrete}
    \centering
    \begin{tabular}{lccc}
    \toprule
    Method & Multi-MNIST & Multi-Fashion & Multi-F+MNIST\\
    \midrule     
    \textbf{SEPNet $\alpha \& \lambda$ } & \textbf{3.04} & \textbf{2.39} & \textbf{2.93}\\
    SEPNet $\alpha$ & 3.02 & 2.37 & 2.92 \\
    SEPNet $\lambda$ & 3.03 & 2.37 & 2.92 \\
    \bottomrule
    \end{tabular}
\end{table}

\begin{figure*}[h]
\centering
\includegraphics[width=1.0\linewidth]{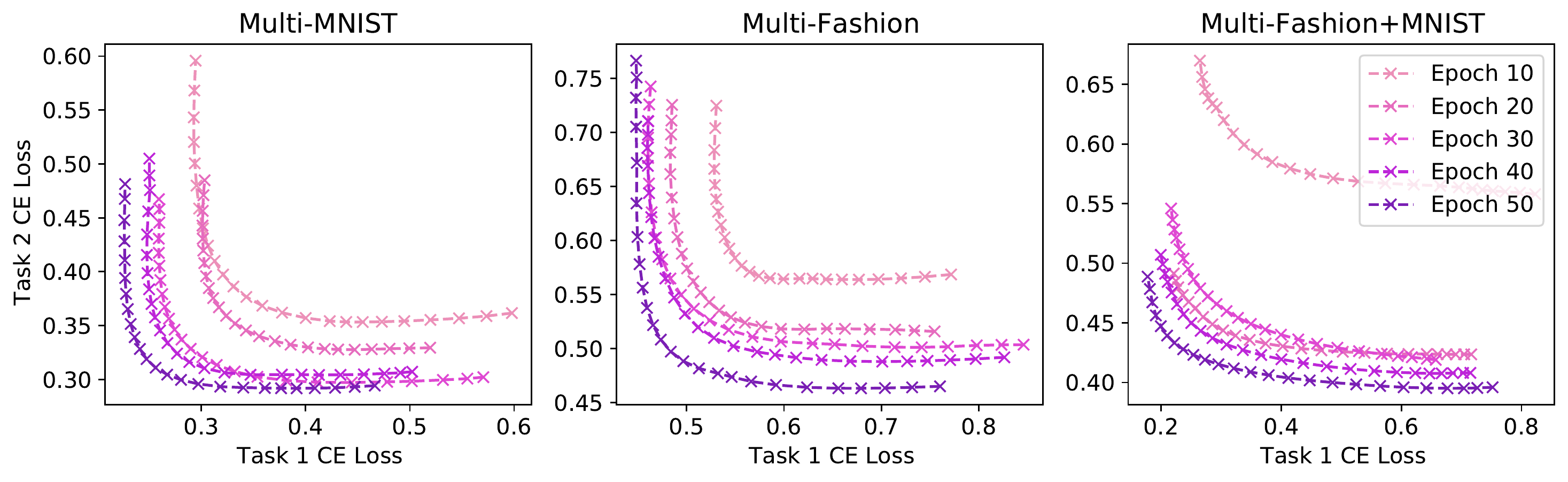}
\vskip -0.1in
\caption{\textbf{SEPNet's convergence graph of the Pareto fronts according to model updates} SEPNet approximates well-spread Pareto fronts from the beginning of training and converges very fast. And, even though only discrete preference rays are used in training due to Sparse Sampling, it can approximate the smooth Pareto front for all preference rays that are not used for training.}
\label{fig:exp_convergence}
\end{figure*}

\begin{figure*}[h]
\centering
\includegraphics[width=1.0\linewidth]{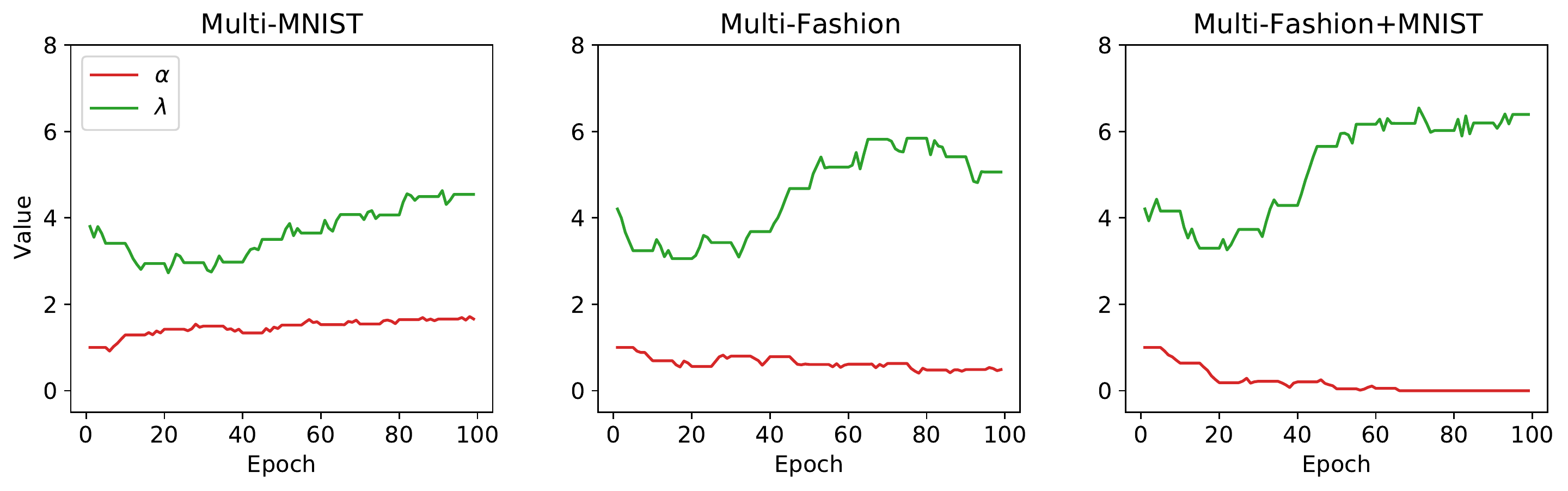}
\vskip -0.1in
\caption{\textbf{Trends in $\alpha$ and $\lambda$ during training for the three Image Classfication datasets} Each parameter is optimized with different optimal points for different datasets.}
\label{fig:exp_convergence}
\end{figure*}

\begin{figure*}[h]
\centering
\includegraphics[width=0.9\linewidth]{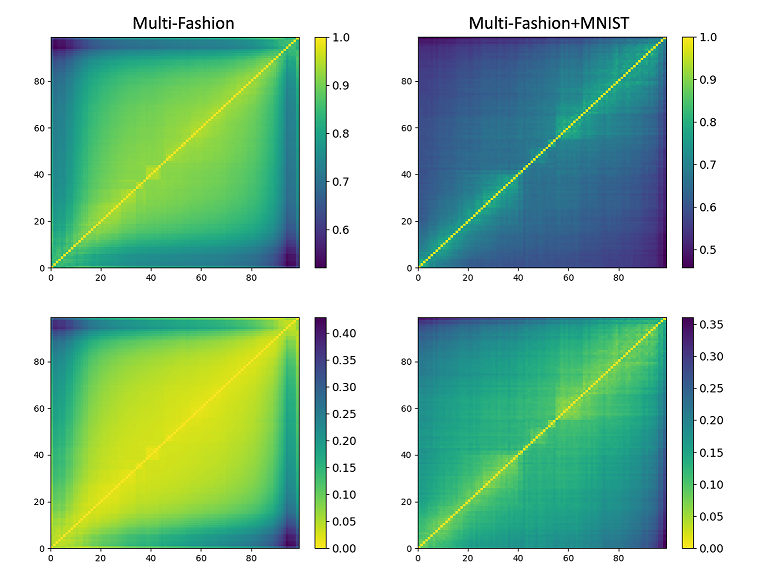}
\caption{Illustration of correlation of preferences. First row uses cosine similarity while the second row uses JS divergence.}
\label{fig:correlation}
\end{figure*}

\begin{table}[h]
\caption{
\textbf{Model size of each method} The separate model requires $n$ models for $n$ preference rays.
}
    \vskip 0.05in
    \label{table:discrete}
    \centering
    \begin{tabular}{lc}
    \toprule
    Method & Number of Parameters\\
    \midrule
     & Multi-MNIST \& Multi-Fashion \& Multi-F+MNIST\\
    \hline
    Single Task & n$\times$42k \\
    ParetoMTL   & n$\times$42k \\
    PHN-LS      & 3,243k \\
    PHN-EPO     & 3,243k \\
    COSMOS      & 43k \\
    SEPNet      & 44k \\
    \midrule
     & Adult \\
    \hline
    Single Task & n$\times$6k \\
    ParetoMTL   & n$\times$6k \\
    PHN-LS      & 716k \\
    PHN-EPO     & 716k \\
    COSMOS      & 7k \\
    SEPNet      & 16k \\
    \midrule
     & Compass\\
    \hline
    Single Task & n$\times$2k \\
    ParetoMTL   & n$\times$2k\\
    PHN-LS      & 304k \\
    PHN-EPO     & 304k \\
    COSMOS      & 2k \\
    SEPNet      & 12k \\
    \midrule
     & Default\\
    \hline
    Single Task & n$\times$7k \\
    ParetoMTL   & n$\times$7k \\
    PHN-LS      & 728k \\
    PHN-EPO     & 728k \\
    COSMOS      & 7k \\
    SEPNet      & 16k \\    
    \bottomrule
    \end{tabular}
\label{table:exp_size}
\end{table}

\end{document}